\documentclass[10pt,twocolumn,letterpaper]{article}

\usepackage{array}
\usepackage{multirow}
\usepackage[pagenumbers]{cvpr} % To force page numbers, e.g. for an arXiv version

\usepackage{float}
\usepackage{placeins}
\definecolor{cvprblue}{rgb}{0.21,0.49,0.74}
\usepackage[pagebackref,breaklinks,colorlinks,allcolors=cvprblue]{hyperref}

\def\paperID{6576} % *** Enter the Paper ID here
\def\confName{CVPR}
\def\confYear{2026}

\title{WildSVG: Towards Reliable SVG Generation \\Under Real-Word Conditions}

\author{
\makebox[\textwidth][c]{%
\begin{minipage}{\textwidth}
\centering
\textbf{Marco Terral}$^{1}$,
\textbf{Haotian Zhang}$^{1,2}$,
\textbf{Tianyang Zhang}$^{1}$,
\textbf{Meng Lin}$^{1}$,
\textbf{Xiaoqing Xie}$^{1}$ \\
\textbf{Haoran Dai}$^{1,3}$,
\textbf{Darsh Kaushik}$^{1,4}$,
\textbf{Pai Peng}$^{1,5}$,
\textbf{Nicklas Scharpff}$^{1}$,
\textbf{David Vazquez}$^{6}$,
\textbf{Joan Rodriguez}$^{1,4}$ \\[0.5em]
$^{1}$QuiverAI,
$^{2}$Columbia University,
$^{3}$Illinois Institute of Technology,
$^{4}$Mila - Quebec Artificial Intelligence Institute,
$^{5}$University of Wisconsin-Madison,
$^{6}$ServiceNow Research
\end{minipage}%
}
}
\begin{document}
\maketitle

\begin{abstract}
We introduce the task of SVG extraction, which consists in translating specific visual inputs from an image into scalable vector graphics. Existing multimodal models achieve strong results when generating SVGs from clean renderings or textual descriptions, but they fall short in real-world scenarios where natural images introduce noise, clutter, and domain shifts. A central challenge in this direction is the lack of suitable benchmarks. To address this need, we introduce the WildSVG Benchmark, formed by two complementary datasets: Natural WildSVG, built from real images containing company logos paired with their SVG annotations, and Synthetic WildSVG, which blends complex SVG renderings into real scenes to simulate difficult conditions. Together, these resources provide the first foundation for systematic benchmarking SVG extraction. We benchmark state-of-the-art multimodal models and find that current approaches perform well below what is needed for reliable SVG extraction in real scenarios. Nonetheless, iterative refinement methods point to a promising path forward, and model capabilities are steadily improving. 
\end{abstract}

\section{Introduction}
Scalable Vector Graphics (SVG) are an XML-based open standard and the leading format for vector graphic representation \citep{svg_creation}, widely adopted in modern image rendering. However, efficiently generating SVGs remains a significant challenge, as the format supports a wide range of primitives, from basic curves such as \textit{path} to more complex shapes like \textit{ellipse} or \textit{polygon}. The task of image vectorization, producing SVG code from rendered images, remains unsolved by the industry. Traditional approaches rely on complex path operations, while deep learning methods struggle to generalize and often underutilize higher-level SVG primitives.\par
\vfill
\begin{figure}[htbp]
        \centering
        \vspace{2cm}
        \includegraphics[width=1\linewidth]{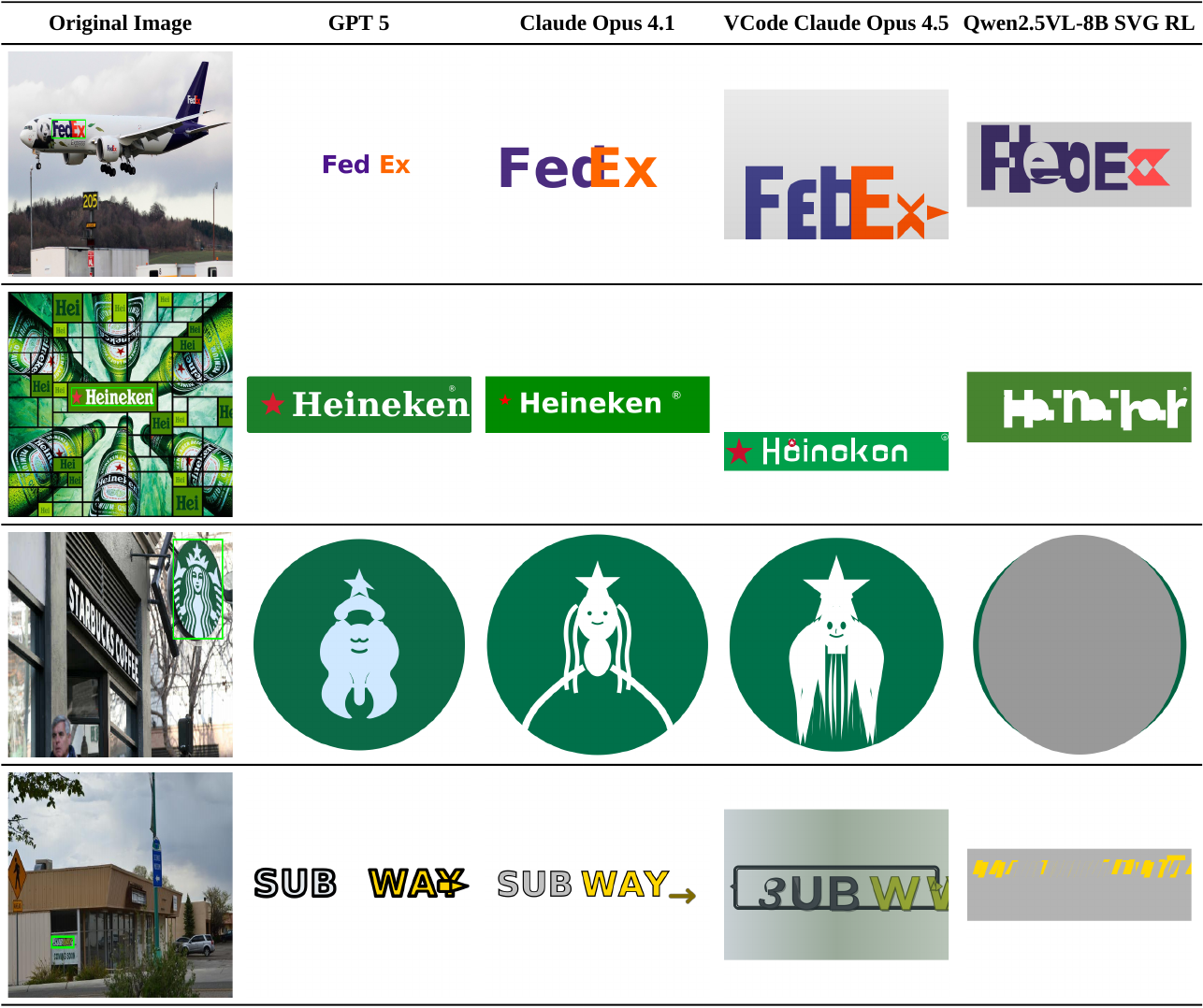}
        \caption{\textbf{Performance on WildSVG}. Extracting the logo from a real image and converting it to SVG using a two step pipeline. Comparison of VLMs on the WildSVG natural dataset.}
        \label{fig:vllm_benchmark_quality_pp_natural}
    \end{figure}

\vfill
    
Recent advances, such as StarVector \citep{starvector}, have demonstrated the potential of multimodal large language models (MLLMs) for SVG generation. StarVector trained on SVG-Stack, dataset of over two million samples, achieved state-of-the-art results through a reinforcement learning pipeline with visual feedback \citep{starvector_rl}. Despite this progress, existing methods are limited to controlled inputs such as clean renderings or text prompts. They fail when confronted with the challenges of natural images, where SVG elements are embedded in cluttered, noisy, and context-rich environments.

\begin{figure*}
    \centering
    \includegraphics[width=1.0\textwidth]{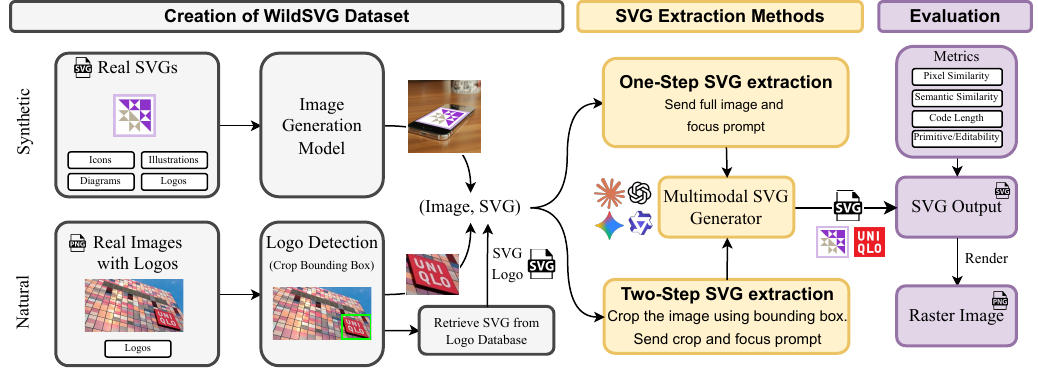}
    \caption{\textbf{Creation of the WildSVG Benchmark and evaluation pipeline}. The dataset combines a synthetic split, built by rendering real SVGs into generated scenes, and a natural split, created by detecting logos in real images and pairing them with SVG annotations. We evaluate SVG extraction using one step and two step multimodal methods and assess outputs through pixel similarity, semantic similarity, code quality, and editability metrics.}
    \label{fig:teaser_diagram}
\end{figure*}
\FloatBarrier

To address this gap, we introduce the SVG extraction task, which focuses on identifying graphical elements such as logos, icons, or pictograms within real-world images, given user guidance, and generating their corresponding SVG code. Unlike full-image vectorization, SVG extraction requires selective abstraction: isolating target elements while ignoring irrelevant visual content; as well as dealing with noises such as textures, shadows, occlusions, and perspective distortions that increase the difficulty of extracting the correct features.\par

We take three key steps to establish SVG extraction as a research problem. 
First, we introduce WildSVG, the first benchmark for this task, composed of two complementary datasets: 
(i) Natural WildSVG, which grounds vector annotations in real-world images, and 
(ii) Synthetic WildSVG, which embeds complex SVGs into natural scenes to simulate challenging visual conditions. 
Second, we define evaluation protocols to support consistent and fair benchmarking across models. 
Finally, we adopt multimodal models, with different training and inference approaches, as baselines, establishing initial performance levels and highlighting open challenges. 
Together, these steps lay the foundation for systematic study of SVG extraction.\par

\section{Related work}

Research on SVG generation is still in its early stages, with most work focused on reconstructing full graphics from clean renderings \citep{starvector}. To the best of our knowledge, no prior work has addressed the SVG extraction task, isolating graphical elements from natural images and generating structured vector representations. Nevertheless, two research directions are closely related and inform our setting: (1) logo detection, and (2) image-to-SVG generation.\par

    %   \begin{figure}[htbp]
    %     \centering
    %     \includegraphics[width=1\linewidth]{figures/synthetic_data.pdf}
    %     \caption{\textbf{Synthetic WildSVG}. Some examples of generated images by the WildSVG Synthetic pipeline.}
    %     \label{fig:syntheticdata_example}
    % \end{figure}
    
\subsection{Logo Detection and Extraction}
Logo detection, a specialized form of object detection, has been widely studied due to applications in multimedia analysis, brand monitoring, and copyright protection. Early approaches relied on hand-crafted features combined with classifiers, while the rise of deep learning established detectors such as YOLO~\citep{yolov11}, DETR~\citep{detr}, and the R-CNN family~\citep{rcnn_family} as the standard. Despite their success, these methods face challenges with dataset imbalance and the closed-set assumption, which limit their ability to generalize to unseen logos~\citep{logo_detection_survey}.\par

Recent work explores zero-shot and open-vocabulary detection to address these issues by leveraging language–vision alignment. For example, some methods replace fixed labels with textual descriptions~\citep{zeroshot_captionopenvocab_obj_detection}, while others combine CLIP-based classifiers with object-agnostic detectors~\citep{zeroshot_logo_detection} or employ transformer-based region embeddings~\citep{zeroshot_simpleopenvocab_obj_detection, gu2022openvocabularyobjectdetectionvision}. Multimodal LLMs~\citep{palix,qwen,molmo_pixmo} further integrate such tasks into pretraining, extending them toward more context-aware object detection~\citep{objdet_llm:context,objdet_llm:rodllm}. However, these approaches output bounding boxes or class labels only, whereas SVG extraction requires both localization and structured vector code generation.\par

\begin{table*}[t]
\caption{\textbf{Overview of publicly available logo detection datasets.} Existing datasets cover a wide range of logo classes, domains and scales, \textit{yet all operate purely in the raster domain}. To the best of our knowledge, \textit{no dataset provides paired SVG annotations}, highlighting the gap that WildSVG aims to fill.}
\label{tab:table_log_det}
\begin{center}
\centering
\renewcommand{\arraystretch}{1.18}
\setlength{\tabcolsep}{10pt}
\begin{tabular}{p{4cm} c c c p{6cm}}
\toprule
\textbf{Name} 
& \textbf{Classes} 
& \textbf{Images} 
& \textbf{Objects} 
& \textbf{Origin and Types} \\
\midrule

\textbf{Belgalogos}~\citep{dataset:belgalogos} 
& 37 
& 10,000 
& 2,695 
& Photojournalist archives, general logos \\

\textbf{FlickrLogos-27}~\citep{dataset:flickrlogos27} 
& 27 
& 1,080 
& 4,671 
& General logos from Flickr search \\

\textbf{FlickrLogos-32}~\citep{dataset:flickrlogos32} 
& 32 
& 2,240 
& 5,644 
& General logos from Flickr search \\

\textbf{SportLogo}~\citep{dataset:sportlogo} 
& 31 
& 2,836 
& --- 
& NHL and NBA logos from web search \\

\textbf{Logos in the Wild}~\citep{dataset:logosinthewild} 
& 871 
& 11,054 
& 32,850 
& Logos captured in diverse real scenes \\

\textbf{QMUL OpenLogo}~\citep{dataset:qmulopenlogo} 
& 352 
& 27,083 
& --- 
& Aggregation of 7 logo detection datasets \\

\textbf{FoodLogoDet-1500}~\citep{dataset:foodlogodet1500} 
& 1,500 
& 99,768 
& 145,400 
& Logos from food industry products \\

\textbf{LogoDet-3K}~\citep{dataset:logodet3k} 
& 3,000 
& 158,652 
& 194,261 
& General logos from large-scale web crawl \\

\bottomrule
\end{tabular}
\end{center}
\end{table*}

\begin{table}[t]
\caption{\textbf{Overview of public SVG generation datasets.} Existing datasets provide large collections of SVGs across icons, fonts and diagrams, but none include natural image contexts or real world SVG extraction settings.}
\label{tab:table_svg_gen}
\centering
\renewcommand{\arraystretch}{1.12}
\setlength{\tabcolsep}{5pt}
\resizebox{\columnwidth}{!}{
\begin{tabular}{p{2.25cm} r r r p{1.9cm} p{2.1cm}}
\toprule
\textbf{Dataset} 
& \textbf{Train} 
& \textbf{Val} 
& \textbf{Test} 
& \textbf{Primitives} 
& \textbf{Annotations} \\
\midrule

\textbf{SVG Stack} 
& 2.1M 
& 108k 
& 5.7k 
& All 
& Captions \\

\textbf{SVG Diagrams} 
& --- 
& --- 
& 472 
& All 
& Captions \\

\textbf{SVG Fonts} 
& 1.8M 
& 91.5k 
& 4.8k 
& Vector paths 
& Font type \\

\textbf{SVG Emoji} 
& 8.7k 
& 667 
& 668 
& All 
& Class \\

\textbf{SVG Icons} 
& 80.4k 
& 6.2k 
& 2.4k 
& Vector paths 
& Class, caption \\

\bottomrule
\end{tabular}
}
\end{table}

\subsection{SVG Generation}
Traditional vectorization methods rely on geometric fitting with the \textit{path} primitive~\citep{iconshop,autotracer,vtracer}, often producing verbose SVG code with limited structural abstraction. Latent-variable models~\citep{vector_fusion,latentsvg1,latentsvg3,latentsvg4} increase flexibility but are typically constrained to narrow SVG subsets and yield non-human-readable outputs.\par

Recent advances expand into specialized domains such as emoji generation \citep{svg_example_dataset1,svg_example_dataset2,svg_example_dataset3}, or employ LLMs for SVG creation and editing \citep{llm_svg_1,llm_svg_2}, framing vector graphics as structured program synthesis \citep{llm_code_1,llm_code_2}. The most significant development is StarVector \citep{starvector}, which casts SVG generation as multimodal inverse rendering and code generation, trained on the large-scale SVG-Stack dataset. A posterior reinforcement learning extension, with rendering feedback (RLRF), further improved its visual fidelity~\citep{starvector_rl}. Yet, these models remain restricted to clean renderings and degrade substantially in natural images with clutter, occlusion, or noise.\par

New studies also present ways for current state-of-the-art VLM (Visual Language Model), like GPT or Claude families, to deal with natural and photographic images for the SVG generation task. More specifically, VCode~\citep{vcode} proposes an iterative analysis and refinement of generated SVG, with usage of visual cues obtained from tools, in order to achieve symbolic SVG representations from images. This approach aims to extract the semantics from images and make a SVG representation faithful to it, rather than generate a high-fidelity SVG.\par

\subsection{Dataset Survey}

The task of identifying and processing SVG renderings within real-world images requires new datasets, \textit{as none currently address this problem}. The StarVector paper~\citep{starvector} introduced SVG-Stack, along with several subsets, as resources for the Image-to-SVG task. While valuable, SVG-Stack focuses on clean renderings and does not capture the challenges of detecting and generating a SVG from natural contexts. Conversely, logo detection datasets provide only localization information (e.g., bounding boxes) without vector annotations.\par

We reviewed existing publicly available logo detection datasets (\cref{tab:table_log_det}). Most provide bounding boxes in natural or semi-natural settings but lack vectorized logo representations. Among them, Logos-in-the-Wild stands out for its scale and diversity, covering difficult real-world conditions such as perspective distortion, scale variation, occlusion, and noisy textures. For SVG generation datasets, we focus on SVG-Stack and its subsets (\cref{tab:table_svg_gen}), which remain the most comprehensive and high-quality resources for vector graphics research. However, they do not include SVGs embedded in real-world image contexts.\par 

Taken together, no existing dataset satisfies the requirements of SVG extraction: grounding vector graphics within natural scenes while maintaining structured SVG annotations. This gap directly motivates the creation of our WildSVG dataset, introduced in the following section.\par

\subsection{Motivation for SVG extraction}

The SVG extraction tasks aim to cover a subset of SVG generation task in which a SVG integrated into an image must be extracted. The idea of this approach is to test models in their capacities to deal with complex noises while discriminating only specific elements. Prior work in logo detection and SVG generation leaves a clear gap as neither tries to achieve this specific task. Detection models can localize target regions but cannot produce structured vector outputs, while SVG generation models excel on clean rasterized SVGs but fail in real-world conditions, where they have to render the demanded features inside challenging scenarios. The SVG extraction task bridges these domains, requiring both localization and vector generation. To enable systematic study of this task, we introduce the WildSVG benchmark, which provides the first datasets designed specifically for SVG extraction.\par

\section{WildSVG Datasets}

To enable systematic study of the SVG extraction task, we introduce the WildSVG datasets, 
consisting of two complementary datasets: 
Natural WildSVG and Synthetic WildSVG. 
Together, they combine the realism of naturally occurring logos with the diversity and controllability of synthetic SVG integration. 
The dataset generation pipelines are illustrated in the Appendix.
%Figures \cref{fig:wildsvg_pipeline_natural} and \cref{fig:wildsvg_pipeline_synthetic}. 

\subsection{Natural WildSVG}
Built from Logos-in-the-Wild \citep{dataset:logosinthewild}, Natural WildSVG augments logo detections 
with vectorized annotations. Each bounding box is paired with (i) an SVG retrieved from worldvectorlogo.com, 
(ii) a textual description, and (iii) a focus prompt specifying the target element. 
To ensure consistency, candidate SVGs were validated using a VLM-based judging model and ranked by DINOv2 features similarity,\citep{dino_score}, 
between rasterized SVGs and cropped detections. 
This process produced high-quality matches between natural logo appearances and their corresponding vector representations.

\subsection{Synthetic WildSVG}
To complement natural logos, Synthetic WildSVG integrates complex SVGs into realistic scenes. 
Starting from SVG-Stack \citep{starvector}, each SVG and its textual description were used to generate synthetic 
compositions with gemini-2.0-flash-preview-image-generation (now known as Nano Banana). 
Prompts were manually optimized to preserve SVG fidelity while embedding the logos naturally in the background.
%(Appendix Fig. \cref{fig:dataset_prompt}). 
We additionally generated focus prompts for the complete dataset and manually annotated bounding boxes for the test split to support reliable evaluation. 
This dataset introduces diverse and complex SVG types under controlled but visually challenging conditions.

\begin{table}[t]
\caption{\textbf{WildSVG} includes \texttt{Natural} samples where real images contain visible logos that are matched to their SVG versions, and \texttt{Synthetic} samples created by using image generation models to place each input SVG into a generated scene. The table reports sample counts, primitives and annotations.}
\label{tab:wildsvg_data}
\centering
\renewcommand{\arraystretch}{1.15}

\resizebox{\columnwidth}{!}{
\begin{tabular}{p{1cm}ccc}
\toprule
\textbf{Dataset} 
& \textbf{Train / Val / Test} 
& \textbf{Primitives} 
& \begin{tabular}[c]{@{}c@{}}\textbf{Annotations}\end{tabular} \\
\midrule

Natural 
& 12759 / 1418 / 227 
& Path 
& \begin{tabular}[c]{@{}p{4.0cm}@{}}Logo brand, focus prompt, description, bounding box\end{tabular} \\

Synthetic 
& 2104 / 190 / 99 
& All 
& \begin{tabular}[c]{@{}p{4.0cm}@{}}Focus prompt, description, bounding box (test only)\end{tabular} \\

\bottomrule
\end{tabular}
}
\end{table}

\begin{figure}[t]
\centering
\includegraphics[width=\linewidth, trim=0 1200 50 0, clip]{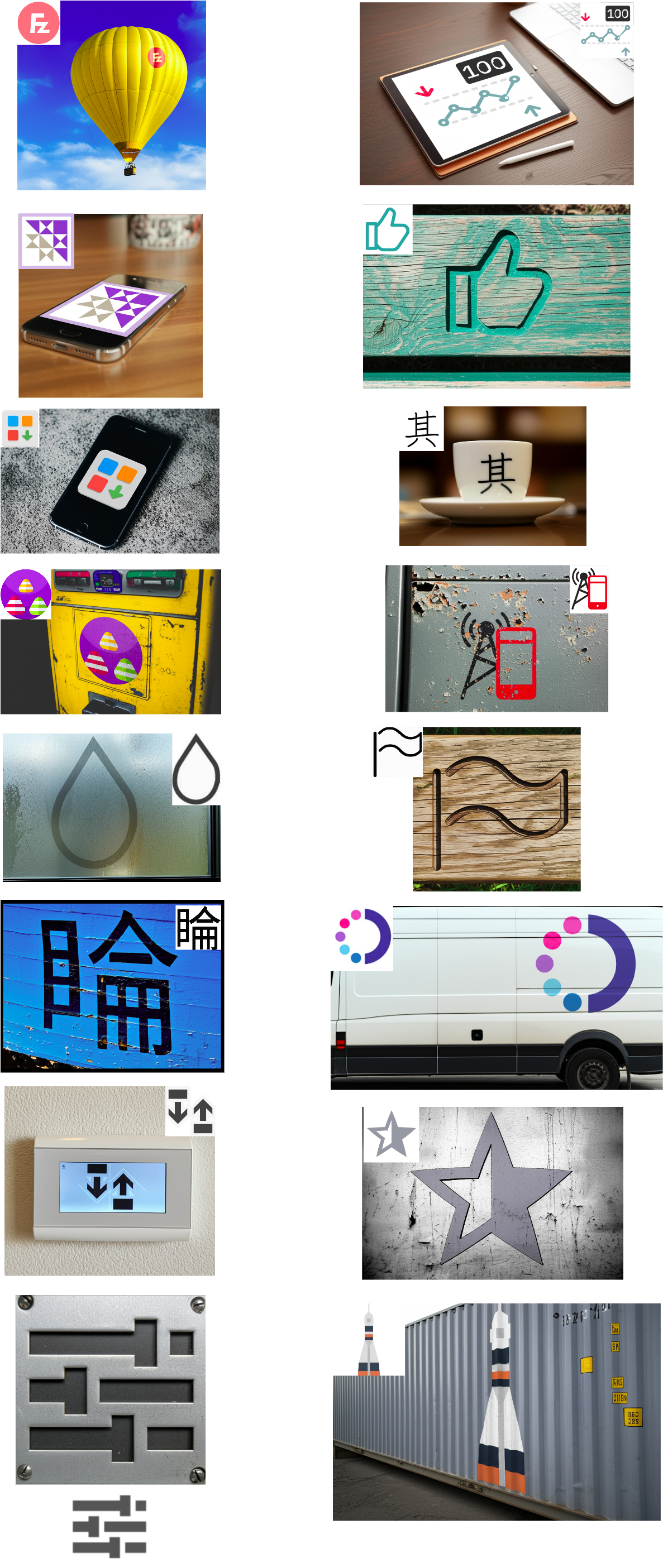}
\caption{\textbf{Examples of Synthetic WildSVG}. Real SVGs extracted from SVG Stack are integrated into realistic scenarios using an image generation model.}
\label{fig:examples_synthetic}
\end{figure}

\begin{figure}[t]
\centering
\includegraphics[width=\linewidth, trim=0 1700 50 0, clip]{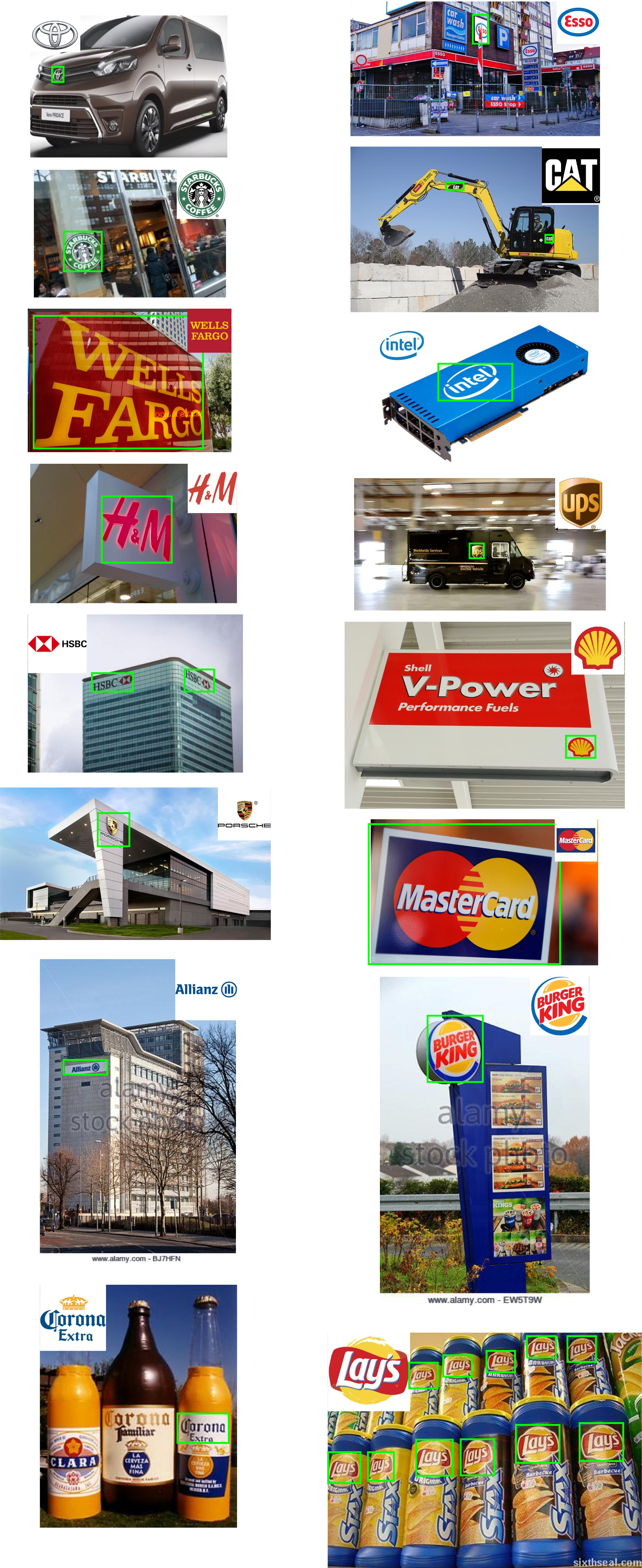}
\caption{\textbf{Examples of Natural WildSVG}. Real images with visible logos are associated with the SVG logos that are extracted from public logo databases.}
\label{fig:examples_natural}
\end{figure}

\subsection{Quality filtering and resulting dataset}
Both datasets underwent automated filtering inspired by StarVector-RL scoring system \citep{starvector_rl}. 
Each sample was scored on \textbf{constancy} (SVG–image similarity), \textbf{alignment} (focus prompt accuracy), 
and, for synthetic data, \textbf{aesthetics} (realism of integration). 
We combined these into aggregate scores and filtered them to at least a 5 out of 10 score, prioritizing constancy and aesthetics over alignment 
(details in Appendix).
%Figures \cref{fig:prompt_judge_natural} and \cref{fig:prompt_judge_synthetic}, and Equation \cref{formula:filter_score}). 
The resulting dataset statistics are shown in \cref{tab:wildsvg_data}, further data about the test split can be found in the Appendix.
%\cref{sec:dataset_analysis}.\par

The final datasets are smaller than SVG-Stack or Logos-in-the-Wild, reflecting their intended use as fine-tuning and evaluation benchmarks rather than pretraining corpora. 
Despite API constraints during synthetic generation, the resulting resources balance natural complexity and synthetic diversity, 
establishing WildSVG as the first benchmark for SVG extraction.

\subsection{Licensing}  
The SVG-Stack data is released under a Creative Commons Attribution 4.0 International (CC BY 4.0) license. 
Accordingly, our Synthetic WildSVG extension is distributed under the same license, allowing both research and commercial use. 
In contrast, due to the copyrighted nature of images employed by Logos-in-the-Wild, 
the Natural WildSVG extension could only be licensed for research purposes.

% Second subfigure
\begin{figure*}[t]
    \centering
    \includegraphics[width=1\linewidth]{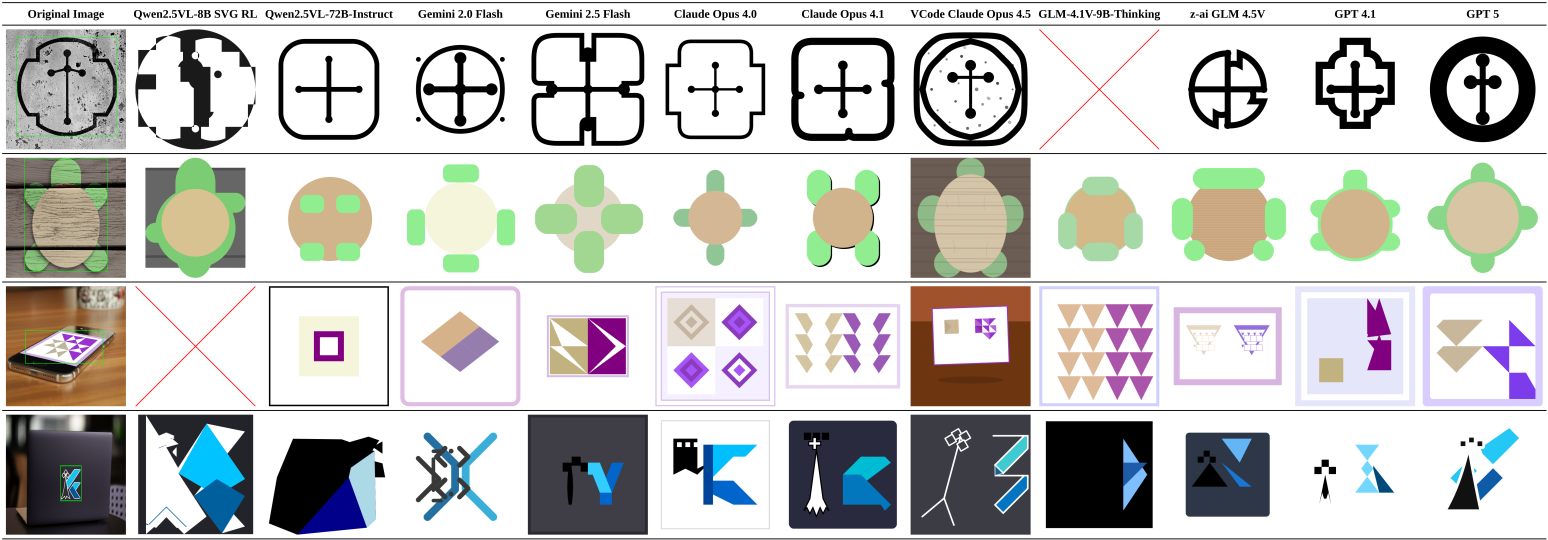}
    \caption{\textbf{Qualitative comparison of VLM outputs on the two step SVG extraction task using the Synthetic split}. Given the cropped logo from the detector, each model produces an SVG reconstruction. Results illustrate varying levels of geometric fidelity, color consistency, and structural correctness across models.}
    \label{fig:complete_vllm_benchmark_quality_pp_synthetic}
\end{figure*}

\section{WildSVG benchmark}

The WildSVG benchmark aims to evaluate model performance on the SVG extraction task in both 
natural scenarios, which involve complex detections and real-world noise, 
and synthetic scenarios, which feature simpler visual noise but more complex SVG structures. 
Employing the test split of each datasets.

\subsection{Evaluation Metrics}
To provide a meaningful assessment of the SVG extraction task, we evaluate models along four dimensions: pixel level fidelity, perceptual and semantic similarity, code compactness and editability. The metrics used are:

\begin{itemize}
    \item \textbf{L2} and \textbf{SSIM} for pixel level fidelity,
    \item \textbf{LPIPS~\citep{lpips}} and \textbf{DINO score~\citep{dino_score}} for perceptual and semantic similarity,
    \item \textbf{Token Length Difference} as a measure of code compactness~\citep{starvector}, computed as the mean absolute difference between the ground truth SVGs and the generated ones to examine how long, or short, the responses are,
    \item \textbf{Primitive Diversity} which we propose as a proxy for SVG editability, since outputs built from a diverse set of simple primitives are easier to manipulate, restyle and modify at design time. It is computed as the average number of unique SVG primitives generated.
\end{itemize}

The use of multiple metrics is motivated by their complementary strengths.  
Pixel level fidelity metrics (L2, SSIM) capture precise reproduction of visual details, although they may penalize outputs that are perceptually faithful but not perfectly aligned at the pixel level. Perceptual and semantic metrics (LPIPS, DINO score) instead capture higher level similarity, complementing fidelity based measures. Code compactness and editability are also important for designers, since shorter and more primitive based SVGs are easier to manipulate and adapt.

% Together, these metrics provide a more comprehensive evaluation, balancing strict accuracy with semantic consistency, alongside SVG code analysis.  

\subsection{Baselines}

We benchmark a broad set of state-of-the-art multimodal model families on the SVG extraction task. The evaluated models include Qwen models (\textbf{Qwen2.5VL 72B Instruct}~\citep{qwen}), Gemini models (\textbf{Gemini 2.0 Flash} and \textbf{Gemini 2.5 Flash}~\citep{comanici2025gemini}), Claude models (\textbf{Claude Opus 4 }and \textbf{Claude Opus 4.1}~\citep{TheC3}), GLM models (\textbf{GLM 4.1V 9B Thinking} and \textbf{GLM 4.5V~\citep{hong2025glm}}), and GPT models (\textbf{GPT 4.1} and \textbf{GPT 5}~\citep{hurst2024gpt}). We also evaluate an SVG specialized model, \textbf{Qwen2.5VL 8B SVG RL} trained with the SFT and RL pipeline introduced in StarVector and RLRF (Reinforcement Learning from Rendering Feedback)~\citep{starvector,starvector_rl}. Finally, we test the a refinement-based method \textbf{VCode}~\citep{vcode} using Claude Sonnet 4.5.

Our goal is to analyze both the performance and trade-offs of each approach, both in one-shot model performance and more sophisticated refinement-based baselines.

% Hopefully add finetuning results too

\subsection{Evaluation Setting}
For a comprehensive evaluation, we conduct SVG extraction under two setups:

\begin{enumerate}
    \item \textbf{Full-image Extraction with Focus Prompt:}  
    The model receives the complete image along with a focus prompt that specifies the target element for extraction.  

    \item \textbf{Two-step Extraction with Perfect Object Localization:}  
    Images are first cropped using perfect bounding boxes before SVG generation.  
    This setting reduces distractions, helping models, particularly StarVector trained models, focus on specific features.  
    It also serves as an upper bound for two-step methods that rely on external detection modules.  
    
\end{enumerate}

\section{WildSVG Benchmark Results}

%%%%%%%%%%%%%%%%%%%% TABLE

\begin{table*}[t]
\caption{\textbf{Main results on WildSVG.} We report visual similarity metrics (DINO, LPIPS, L2, SSIM) and code metrics (length difference and primitive diversity) for one-step and two-step SVG extraction on the Natural and Synthetic splits. Lower is better for LPIPS, L2, and length difference. Higher is better for DINO, SSIM, and primitive diversity.}
\label{tab:unified_svg_extraction}
\centering

\resizebox{\textwidth}{!}{
\begin{tabular}{lcccccccccccc}
\toprule
\multicolumn{13}{c}{\large \textbf{One-step generation}} \\
\midrule

\multirow{2}{*}{\textbf{Model}} 
& \multicolumn{6}{c}{\textbf{Natural}} 
& \multicolumn{6}{c}{\textbf{Synthetic}} \\
\cmidrule(lr){2-7} \cmidrule(lr){8-13}
& \textbf{DINO ↑} & \textbf{LPIPS ↓} & \textbf{L2 ↓} & \textbf{SSIM ↑}
& \textbf{Len diff ↓} & \textbf{Prim div ↑}
& \textbf{DINO ↑} & \textbf{LPIPS ↓} & \textbf{L2 ↓} & \textbf{SSIM ↑}
& \textbf{Len diff ↓} & \textbf{Prim div ↑} \\
\midrule

Qwen2.5VL 8B SVG RL
& 0.69 & \underline{0.39} & \textbf{0.15} & \textbf{0.63} & 8108 & 2.57
& 0.76 & 0.43 & \underline{0.16} & \underline{0.61} & 10375 & \textbf{3.12} \\

Qwen2.5VL 72B Instruct  
& 0.77 & 0.41 & 0.22 & 0.58 & 8804 & 2.15
& 0.77 & 0.42 & 0.21 & 0.58 & 1516 & 1.86 \\

GLM 4.5V
& \underline{0.79} & 0.39 & 0.18 & 0.61 & 8710 & 2.02
& 0.77 & \underline{0.40} & 0.19 & 0.59 & 1371 & 1.95 \\

VCode Claude Sonnet 4.5
& 0.78 & 0.46 & 0.22 & 0.53 & 8141 & \underline{3.34}
& \textbf{0.85} & \textbf{0.38} & \textbf{0.20} & \textbf{0.59} & \underline{920} & 2.66 \\

Gemini Flash 2.5
& 0.79 & 0.42 & 0.20 & 0.58 & \underline{8509} & 2.11
& 0.78 & 0.43 & 0.21 & 0.57 & \textbf{193} & 1.82 \\

Claude Opus 4.1
& \textbf{0.80} & \textbf{0.40} & \underline{0.19} & \underline{0.61} & 8919 & 2.07
& \underline{0.80} & \underline{0.42} & 0.20 & 0.58 & 1441 & 2.03 \\

GPT 5
& \textbf{0.80} & 0.40 & \underline{0.19} & 0.58 & \textbf{6815} & \textbf{2.54}
& \underline{0.79} & 0.42 & 0.22 & 0.57 & 631 & \underline{2.21} \\

\midrule
\midrule
\multicolumn{13}{c}{\large \textbf{Two-step generation}} \\
\midrule

\multirow{2}{*}{\textbf{Model}} 
& \multicolumn{6}{c}{\textbf{Natural}} 
& \multicolumn{6}{c}{\textbf{Synthetic}} \\
\cmidrule(lr){2-7} \cmidrule(lr){8-13}
& \textbf{DINO ↑} & \textbf{LPIPS ↓} & \textbf{L2 ↓} & \textbf{SSIM ↑}
& \textbf{Len diff ↓} & \textbf{Prim div ↑}
& \textbf{DINO ↑} & \textbf{LPIPS ↓} & \textbf{L2 ↓} & \textbf{SSIM ↑}
& \textbf{Len diff ↓} & \textbf{Prim div ↑} \\
\midrule

Qwen2.5VL 8B SVG RL
& 0.74 & 0.46 & \underline{0.18} & 0.60 & \underline{1323} & \textbf{3.05}
& 0.82 & 0.37 & \underline{0.16} & 0.63 & 3063 & \textbf{3.20} \\

VCode Claude Sonnet 4.5
& 0.77 & 0.58 & 0.30 & 0.45 & 7768 & \underline{3.04}
& 0.87 & 0.44 & 0.22 & 0.55 & \underline{690} & \underline{2.90} \\

Qwen2.5VL 72B Instruct
& 0.81 & 0.36 & 0.21 & 0.62 & 8337 & 1.86
& 0.85 & 0.34 & 0.20 & 0.61 & 1484 & 1.99 \\

GLM 4.5V
& 0.83 & 0.34 & 0.20 & 0.63 & 8302 & 1.98
& 0.86 & \underline{0.32} & 0.18 & \textbf{0.64} & 1387 & 1.97 \\

Gemini Flash 2.5
& 0.85 & \underline{0.32} & 0.19 & \underline{0.64} & 8669 & 2.06
& 0.88 & 0.33 & 0.19 & \underline{0.64} & 1074 & 1.86 \\

Claude Opus 4.1
& \underline{0.86} & \textbf{0.32} & \textbf{0.16} & \textbf{0.66} & 8798 & 2.04
& \textbf{0.90} & \textbf{0.30} & \textbf{0.16} & \textbf{0.65} & 1405 & 2.00 \\

GPT 5
& \textbf{0.87} & 0.34 & \textbf{0.18} & \textbf{0.63} & \textbf{6448} & 2.30
& \underline{0.89} & \underline{0.31} & \underline{0.18} & \underline{0.63} & \textbf{921} & 2.13 \\

\bottomrule
\end{tabular}
}
\end{table*}

%%%%%%%%%%%%%%%%%%%%%%%%%

From our current benchmark, we present reduced tables, containing the most recent model of each family; the complete results are provided in the Appendix.%\cref{sec:complete_benchmark}.
Since models within a family generally exhibit similar behavior, we focus on the most recent and best-performing representatives. We can also observe more qualitative results in the same previous section.\par

\begin{figure}[t]
        \centering
        \includegraphics[width=1\linewidth]{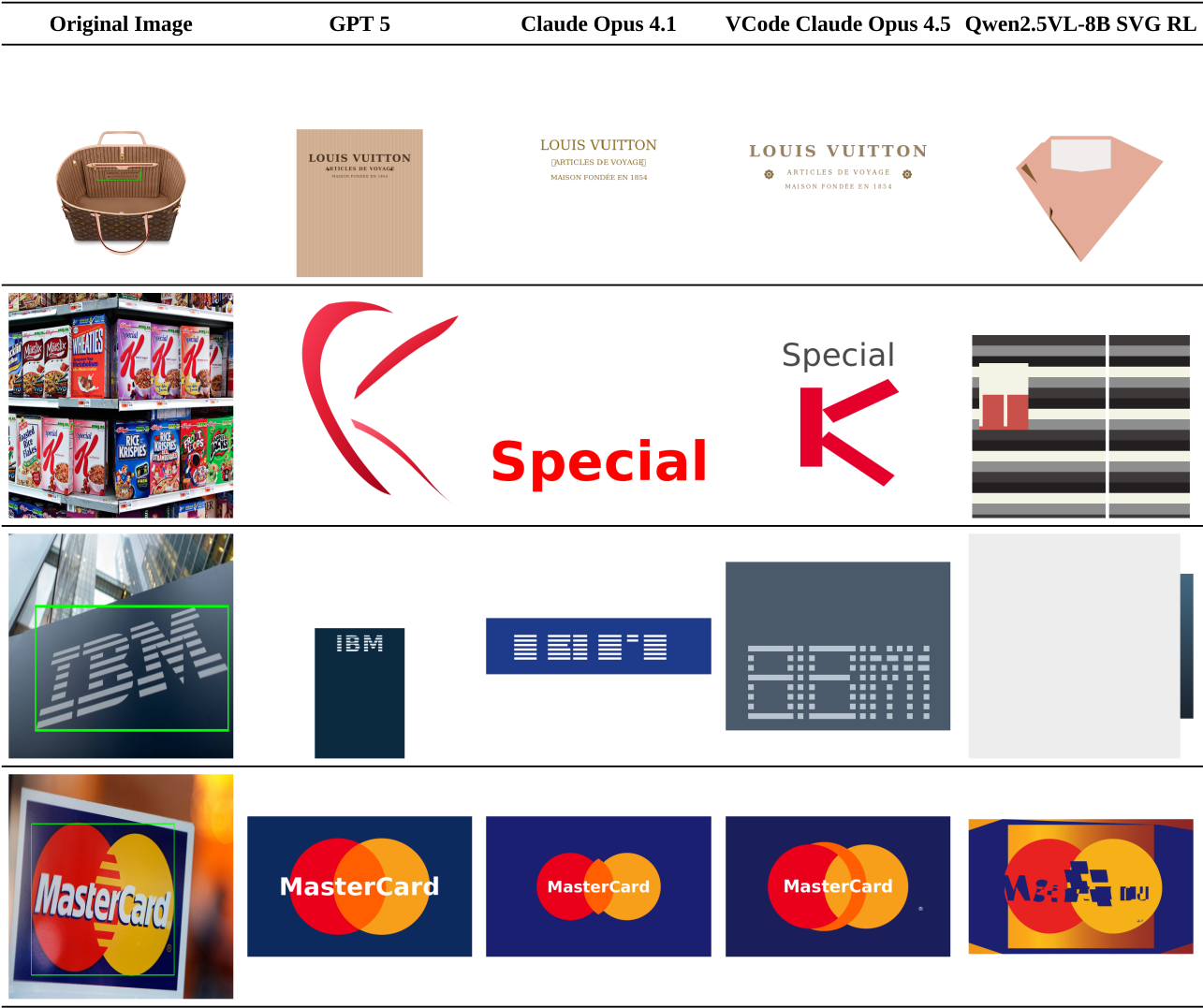}
        \caption{Comparison of VLMs for one-step SVG extraction natural dataset}
        \label{fig:vllm_benchmark_quality_full_natural}
    \end{figure}

\begin{figure}[t]
        \centering
        \includegraphics[width=1\linewidth]{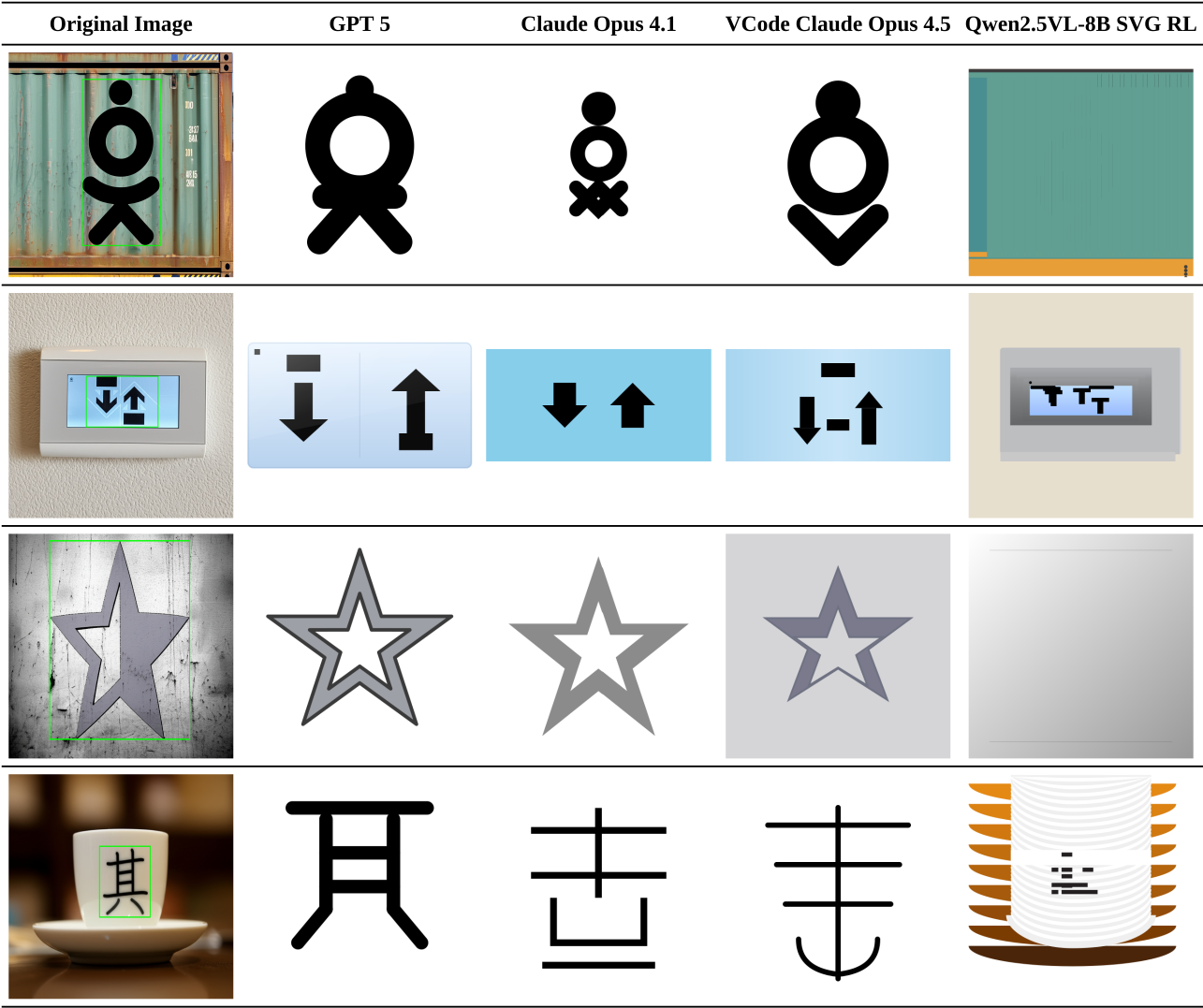}
        \caption{Comparison of VLMs for one-step SVG extraction synthetic dataset}
        \label{fig:vllm_benchmark_quality_full_synthetic}
    \end{figure}

Two clear trends emerge from the results. First, across families, models produce SVGs that follow broadly similar patterns, with only a few notable outliers. Some outliers like Qwen2.5VL-8B RL SVG, frequently attempt to render the entire image rather than isolating the SVG. Other specific cases are the VCode inference, an approach used with Claude Sonnet 4.5, where we can see it render the background or some minor details, like the Star SVG in \cref{fig:vllm_benchmark_quality_full_synthetic} or the Subway logo in \cref{fig:vllm_benchmark_quality_pp_natural}. Second, models consistently achieve higher scores on the synthetic dataset, reflecting the greater difficulty of the natural dataset (more information in the Appendix);
%\cref{sec:dataset_analysis})%;
where scaling, perspective distortion, occlusions, shadows, and noise complicate vectorization. As shown in \cref{fig:vllm_benchmark_quality_full_natural}, even advanced models struggle with ambiguous cases such as the Special K box, where only GPT 5 and VCode approach capture the “K” logo.\par

Observing \cref{tab:unified_svg_extraction}, we can conclude that detection itself does not consistently improve results across families. The notable exception is Qwen2.5VL-8B SVG RL, trained with Starvector pipeline, which in the one-step setting ignores the prompt and attempts to vectorize the full image (\cref{fig:vllm_benchmark_quality_full_natural}). Despite the fact than other VLM families, including Qwen, do have the capacity to focus on the given prompt. This behavior suggests a weakness in StarVector’s training pipeline, as text-to-SVG and image-to-SVG tasks are learned separately in this regime which may reduce the alignment between prompt and image features during SVG generation.\par

    % First subfigure

    % Second subfigure

\begin{figure}[t]
        \centering
        \includegraphics[width=1\linewidth]{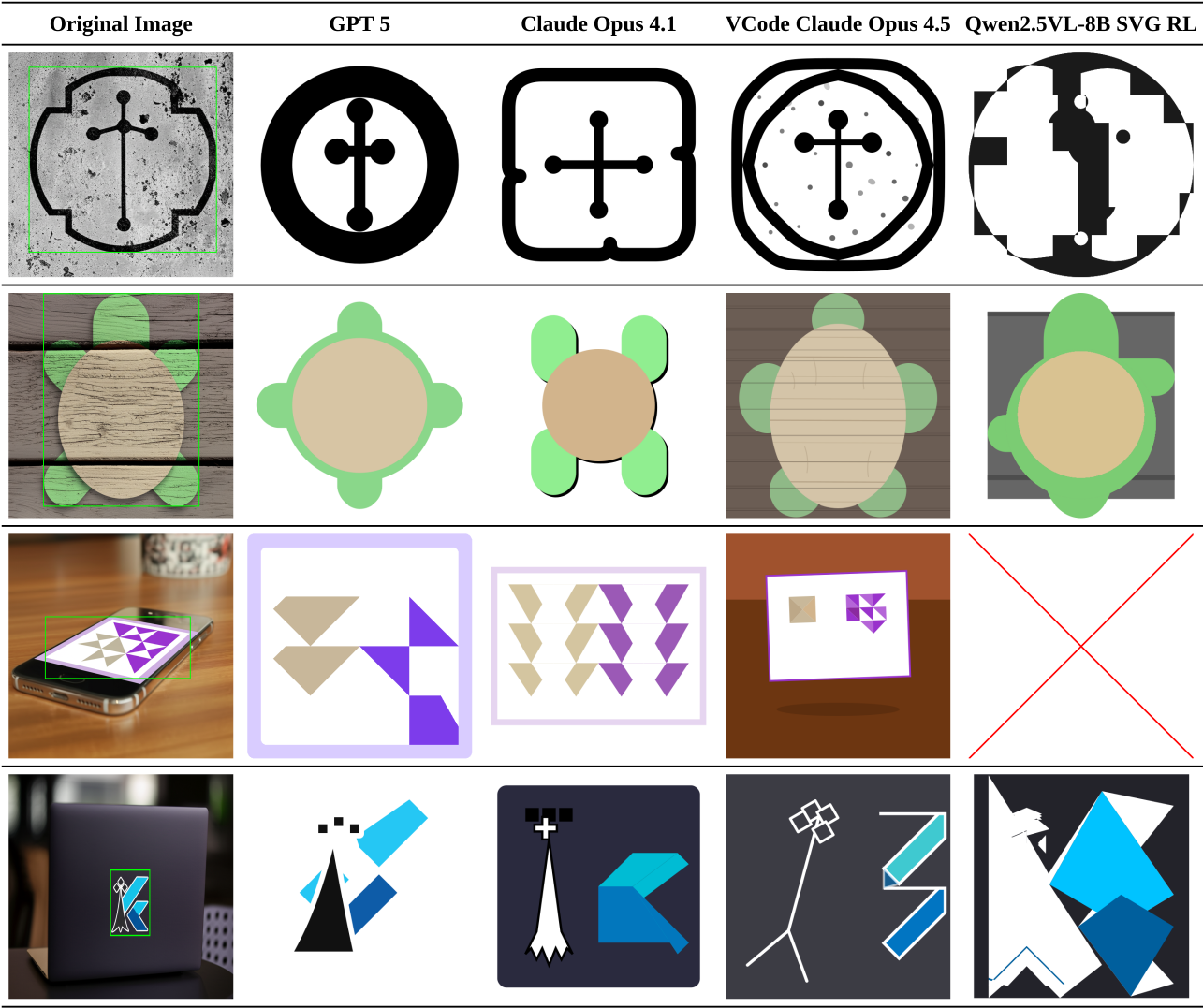}
        \caption{Comparison of VLMs for two-step SVG extraction synthetic dataset}
        \label{fig:vllm_benchmark_quality_pp_synthetic}
    \end{figure}

Across families, \textbf{VLMs generally optimize for semantic similarity rather than aesthetic fidelity}. This tendency is reflected in the frequent use of the \textit{text} primitive to approximate letters with similar fonts, rather than rendering them as shapes. As a result, SVGs achieve strong performance on semantic metrics (LPIPS, DINO) but weaker performance on pixel-level metrics (L2, SSIM). For example, the rendering of the Heineken logo in the GPT 5 model (\cref{fig:vllm_benchmark_quality_pp_natural}) appears convincing in the overall structure but reveals clear inaccuracies upon closer examination. Synthetic examples further highlight this trade-off: some models reproduce SVGs with structures reminiscent of the original, but insufficiently faithful for precise extraction (\cref{fig:vllm_benchmark_quality_full_synthetic}, \cref{fig:vllm_benchmark_quality_pp_synthetic}). This semantic approach generates simpler and shorter SVGs; in fact, the most accurate ones tend to be longer near the GT length, thus allowing for better expressivity. Despite Claude 4.1 and GPT 5 delivering the most semantically consistent and highest-fidelity SVG, both remain below the required threshold for a complete solution of SVG extraction.

Qwen2.5VL-8B SVG RL and VCode diverge from this general trend, sacrificing semantic fidelity in favor of visual aesthetics. This is reflected in the metrics, where L2 and SSIM are prioritized over LPIPS and DINO, and in qualitative examples such as the FedEx and Heineken logos (\cref{fig:vllm_benchmark_quality_pp_natural}). In these cases, they often rely on shape primitives to render letters, a strategy preferable to text primitives since reproducing exact font styles, kerning, and spacing is effectively impossible. While this approach reduces semantic scores, it has the potential to produce SVGs visually closer to the original designs. Another effect of this approach is that the generated SVGs tend also to be longer and more expressive with primitives (to the detriment of Qwen2.5VL-8B SVG RL and improving VCode length metrics) ; this make intuitive sense, employing more primitives might help achieve complex structures but is expensive with tokens. However, \textit{for the specific task of SVG extraction, Qwen2.5VL-8B SVG RL’s results remain below the semantic and aesthetic quality achieved by Claude and GPT models}, even when solving its alignment problem by providing a detected SVG. We hypothesize that noise in real-world scenarios, such as textures and shadows, may overwhelm the model, leading it to overfit to subtle visual variations rather than isolating the core SVG structures. And thus explaining why this model still shows strong capabilities in generating complex code directly from rasterized SVGs. However the VCode approach shows greater potential, providing similar SVG extraction approaches while improving on its failures. With VCode inference we are able to increase the performance of Claude Sonnet 4.5, \cref{tab:unified_svg_extraction}, while maintaining the focusing capacities of the model. We can also observe the same philosophy of SVGs with greater potential for accuracy while providing more semantical readability. On top of that theses SVGs aren't as long while maintaining the primitive diversity.\par

Overall, our baseline demonstrates that current VLMs can generate simpler SVGs that are semantically meaningful but still fall short in aesthetic fidelity and primitive potential. Across families, most models achieve relatively similar scores regardless of the visual encoder or LLM architecture, suggesting that model size and inference technique play a greater role than design choices. As Open-source models, typically ranging from 10–70B parameters, tend to perform slightly worse than larger proprietary systems. However, even the strongest models, such as Claude 4.1 and GPT 5, plateau at approximately DINO 90, LPIPS 30, SSIM 60, and L2 15. Even employing more complex inference approaches like VCode. By comparison, achieving high-fidelity SVG generation would require scores closer to DINO 95, LPIPS 10, SSIM 80, and L2 9. In fact, our code metrics indicate an under usage of tokens and primitives, clearly illustrated in the qualitative results. These findings point to a performance ceiling in current approaches which still require to be addressed.\par

    % Second subfigure

\section{Conclusion}

We introduced the task of SVG extraction, extending multimodal models to generate vector graphics directly from natural images, and proposed WildSVG, the first benchmark for this problem. WildSVG combines real-world logos with synthetic compositions, enabling evaluation under both natural and controlled conditions.\par  

Our benchmarking of leading VLM families reveals three consistent takeaways:  
(1) models perform better on synthetic than natural data, showing the impact of real-world distortions;  
(2) current systems prioritize semantic similarity over pixel fidelity;  
(3) even the strongest models plateau below high-fidelity thresholds, leaving clear headroom for improvement. While Claude and GPT balance fidelity and semantics most effectively, Qwen2.5VL-8B SVG RL highlights a contrasting trade-off by favoring aesthetics over semantics while still lacking to achieve competitive performance. Iterative inference show promising results if combined with a strong VLM.\par 

Looking ahead, we identify several open research directions: (i) improving alignment between prompts and structured vector outputs, particularly for StarVector training; (ii) integrating SVG generation and extraction tasks into VLM training pipelines to improve fidelity in vector code; (iii) extending two-step approaches to leverage SVG generation without requiring task-specific fine-tuning; (iiii) study the impact of iterative approaches in better trained model; and (iiiii) expanding WildSVG datasets to allow broader training approaches not only finetuning.\par

By framing SVG extraction as a benchmarked task, we aim to catalyze future research at the intersection of vision, language, and structured graphics generation. \par 

{
   \small
   \bibliographystyle{ieeenat_fullname}
   \bibliography{main}
}

\clearpage
\setcounter{page}{1}
\maketitlesupplementary

\section{Appendix}
\subsection{LLM usage}

Parts of this manuscript were refined and polished using ChatGPT (GPT-5), a large language model developed by OpenAI. The model was employed solely for language editing and clarity improvements; all technical content, data analyses, and conceptual contributions remain the original work of the authors.\par

\subsection{Complete benchmark}
\label{sec:complete_benchmark}
In ~\cref{tab:complete_svg_extraction_benchmark_unified} we report the benchmark results for all VLMs evaluated in our study.\par

\begin{table*}[ht!]
\caption{\textbf{Complete unified results on WildSVG.} We include the diverse families and versions benchmarked on WildSVG, including both the one-step and two-step SVG extraction on the Natural and Synthetic splits.}
\label{tab:complete_svg_extraction_benchmark_unified}
\begin{center}
\centering
\renewcommand{\arraystretch}{1.2}
\setlength{\tabcolsep}{4pt}

\resizebox{\textwidth}{!}{
\begin{tabular}{lcccccccccccc}
\toprule
\multicolumn{13}{c}{\large \textbf{One-step generation}} \\
\midrule

\multirow{2}{*}{\textbf{Model}} 
& \multicolumn{6}{c}{\textbf{Natural}} 
& \multicolumn{6}{c}{\textbf{Synthetic}} \\
\cmidrule(lr){2-7} \cmidrule(lr){8-13}
& \textbf{DINO ↑} & \textbf{LPIPS ↓} & \textbf{L2 ↓} & \textbf{SSIM ↑}
& \textbf{Len diff ↓} & \textbf{Prim div ↑}
& \textbf{DINO ↑} & \textbf{LPIPS ↓} & \textbf{L2 ↓} & \textbf{SSIM ↑}
& \textbf{Len diff ↓} & \textbf{Prim div ↑} \\
\midrule

Qwen2.5VL-72B-Instruct
& 0.77 & 0.41 & 0.22 & 0.58 & 8803.83 & 2.15
& 0.77 & 0.42 & 0.21 & 0.58 & 1516.36 & 1.86 \\

Qwen2.5VL-8B SVG RL
& 0.69 & 0.39 & \textbf{0.15} & \textbf{0.63} & \underline{8108.38} & \underline{2.57}
& 0.76 & 0.43 & \textbf{0.16} & 0.61 & 10374.72 & 3.12 \\

Gemini Flash 2
& 0.78 & \underline{0.38} & \underline{0.17}  & \underline{0.61} & 8233.99 & 1.69
& 0.83 & \textbf{0.32} & 0.19 & \underline{0.63} & 1202.83 & 1.51 \\

Gemini Flash 2.5
& 0.79 & 0.42 & 0.20 & 0.58 & 8509.60 & 2.11
& 0.78 & 0.43 & 0.21 & 0.57 & 192.90 & 1.82 \\

Claude Opus 4
& 0.78 & 0.40 & 0.18 & 0.60 & 8895.15 & 2.35
& 0.84 & \underline{0.34} & 0.19 & 0.62 & 1395.59 & 2.17 \\

Claude Opus 4.1
& \underline{0.80} & 0.40 & 0.19 & \underline{0.61} & 8919.63 & 2.07
& 0.80 & 0.42 & 0.20 & 0.58 & 1441.08 & 2.03 \\

VCode Claude Sonnet 4.5
& 0.78 & 0.46 & 0.22 & 0.53 & 8140.55 & \textbf{3.34}
& \underline{0.85} & 0.38 & 0.20 & 0.59 & 920.07 & 2.66 \\

GLM-4.1V-9B-Thinking
& 0.75 & \textbf{0.37} & \underline{0.17} & \textbf{0.63} & 8954.52 & 2.13
& 0.78 & 0.37 & 0.21 & 0.62 & 1295.58 & 1.87 \\

z-ai GLM 4.5V
& 0.79 & 0.39 & 0.18 & \underline{0.61} & 8710.47 & 2.02
& 0.77 & 0.40 & 0.19 & 0.59 & 1370.70 & 1.95 \\

GPT 4.1
& \textbf{0.81} & 0.39 & 0.18 & 0.59 & 8775.49 & 2.18
& \textbf{0.86} & \textbf{0.32} & \underline{0.18} & \textbf{0.64} & 1237.15 & 1.93 \\

GPT 5
& \underline{0.80} & 0.40 & 0.19 & 0.58 & \textbf{6814.91} & 2.54
& 0.79 & 0.42 & 0.22 & 0.57 & 630.88 & 2.21 \\

\midrule
\midrule
\multicolumn{13}{c}{\large \textbf{Two-step generation}} \\
\midrule

\multirow{2}{*}{\textbf{Model}} 
& \multicolumn{6}{c}{\textbf{Natural}} 
& \multicolumn{6}{c}{\textbf{Synthetic}} \\
\cmidrule(lr){2-7} \cmidrule(lr){8-13}
& \textbf{DINO ↑} & \textbf{LPIPS ↓} & \textbf{L2 ↓} & \textbf{SSIM ↑}
& \textbf{Len diff ↓} & \textbf{Prim div ↑}
& \textbf{DINO ↑} & \textbf{LPIPS ↓} & \textbf{L2 ↓} & \textbf{SSIM ↑}
& \textbf{Len diff ↓} & \textbf{Prim div ↑} \\
\midrule

Qwen2.5VL-72B-Instruct
& 0.81 & 0.36 & 0.21 & 0.62 & 8337.19 & 1.86
& 0.85 & 0.34 & 0.20 & 0.61 & 1483.57 & 1.99 \\

Qwen2.5VL-8B SVG RL
& 0.74 & 0.46 & 0.18 & 0.60 & 1323.05 & 3.05
& 0.82 & 0.37 & 0.16 & 0.63 & 3062.99 & 3.20 \\

Gemini Flash 2
& 0.76 & 0.40 & 0.18 & 0.61 & 8328.87 & 1.58
& 0.86 & 0.32 & 0.18 & 0.64 & 1307.80 & 1.65 \\

Gemini Flash 2.5
& 0.85 & 0.32 & 0.19 & 0.64 & 8668.80 & 2.06
& 0.88 & 0.33 & 0.19 & 0.64 & 1074.12 & 1.86 \\

Claude Opus 4
& 0.78 & 0.43 & 0.19 & 0.58 & 8647.89 & 2.16
& 0.88 & 0.30 & 0.15 & 0.66 & 1352.99 & 2.08 \\

Claude Opus 4.1
& 0.86 & 0.32 & 0.16 & 0.66 & 8797.66 & 2.04
& 0.90 & 0.30 & 0.16 & 0.65 & 1405.42 & 2.00 \\

VCode Claude Sonnet 4.5
& 0.77 & 0.58 & 0.30 & 0.45 & 7767.66 & 3.04
& 0.87 & 0.44 & 0.22 & 0.55 & 689.90 & 2.90 \\

GLM-4.1V-9B-Thinking
& 0.76 & 0.41 & 0.20 & 0.59 & 8488.48 & 1.89
& 0.83 & 0.33 & 0.18 & 0.63 & 1327.17 & 1.90 \\

z-ai GLM 4.5V
& 0.83 & 0.34 & 0.20 & 0.63 & 8301.63 & 1.98
& 0.86 & 0.32 & 0.18 & 0.64 & 1386.77 & 1.97 \\

GPT 4.1
& 0.80 & 0.41 & 0.20 & 0.57 & 8779.01 & 2.01
& 0.88 & 0.31 & 0.17 & 0.63 & 1475.92 & 2.04 \\

GPT 5
& 0.87 & 0.34 & 0.18 & 0.63 & 6448.23 & 2.30
& 0.89 & 0.31 & 0.18 & 0.63 & 921.46 & 2.13 \\

\bottomrule
\end{tabular}
}
\end{center}
\end{table*}

\subsection{Dataset generation}

\subsubsection{Dataset generation pipelines}
\begin{figure*}[htbp]
    \centering
    \includegraphics[width=1\linewidth]{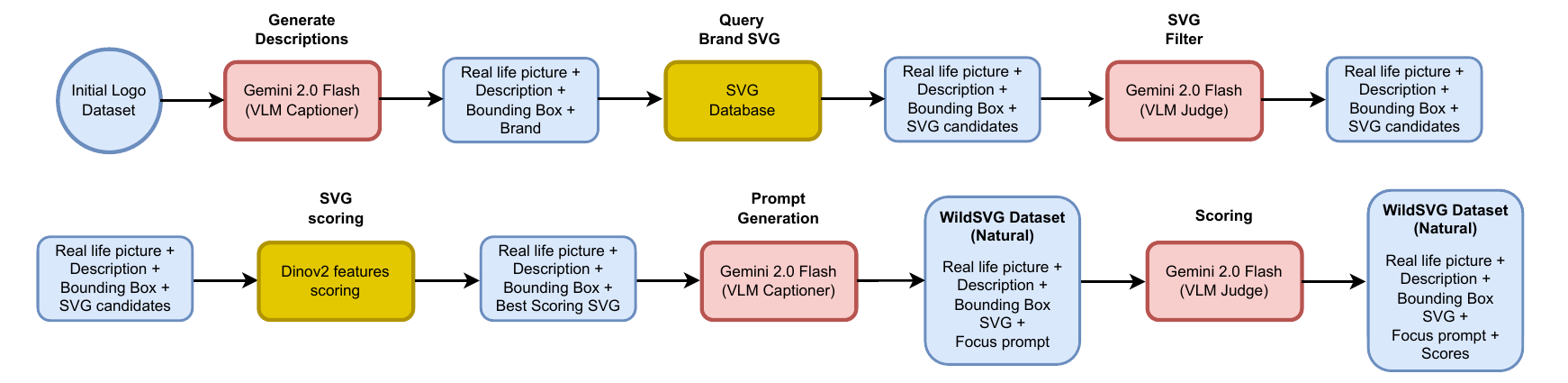}
    \caption{Pipeline for synthetic WildSVG generation}
    \label{fig:wildsvg_pipeline_natural}
\end{figure*}

\begin{figure*}[htbp]
    \centering
    \includegraphics[width=0.9\linewidth]{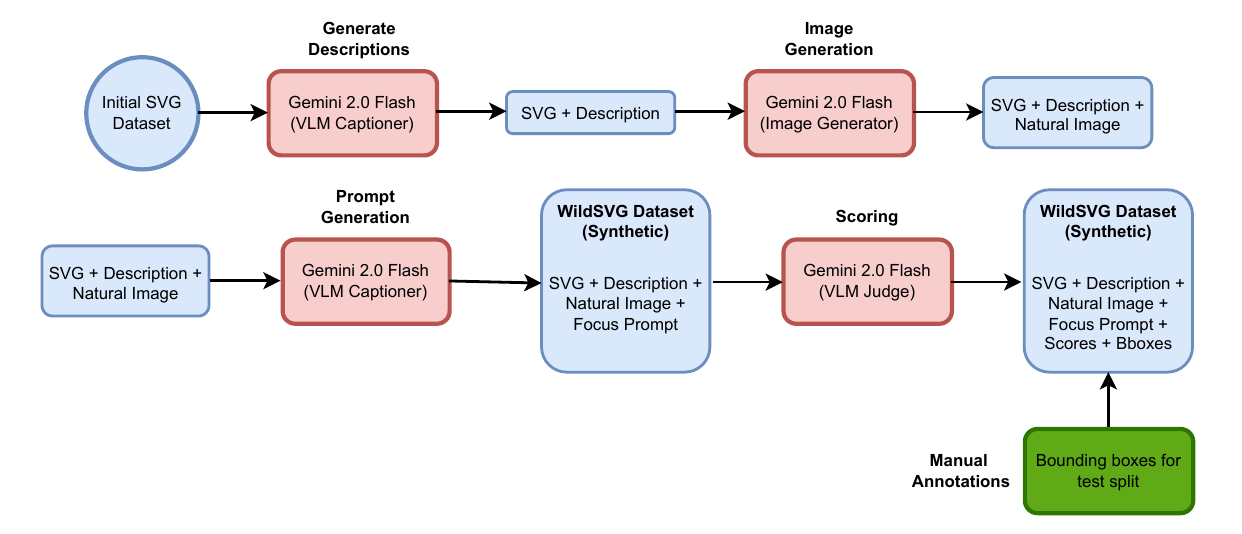}
    \caption{Pipeline for natural WildSVG generation}
    \label{fig:wildsvg_pipeline_synthetic}
\end{figure*}

\begin{figure}[ht!]
    \centering
    \includegraphics[width=0.9\linewidth]{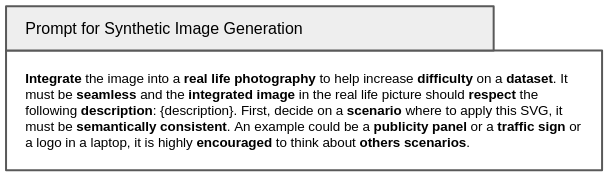}
    \caption{Prompt for image generation of Synthetic WildSVG Dataset}
    \label{fig:dataset_prompt}
\end{figure}

We present in \cref{fig:wildsvg_pipeline_natural} and \cref{fig:wildsvg_pipeline_synthetic} the processing pipelines employed for generating the WildSVG datasets. In \cref{fig:dataset_prompt} we can observe the prompt employed in the synthetic WildSVG dataset pipeline for integrating SVGs into images. This prompt was generated manually after multiple iterations, achieving the most aesthetic pleasing results while maintaining most or all the features of the original SVG.\par

\begin{figure}[ht!]
        \centering
        \includegraphics[width=0.8\linewidth]{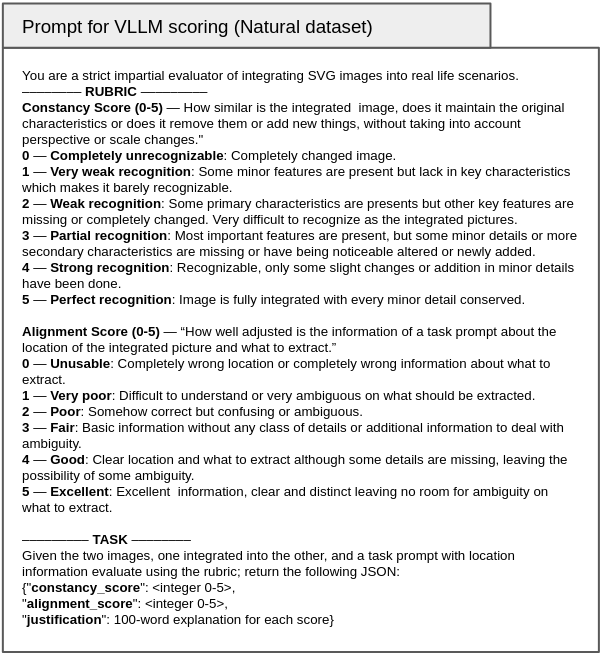}
        \caption{Prompt for scoring Natural WildSVG instances}
        \label{fig:prompt_judge_natural}
\end{figure}

\begin{figure}[ht!]
        \centering
        \includegraphics[width=0.8\linewidth]{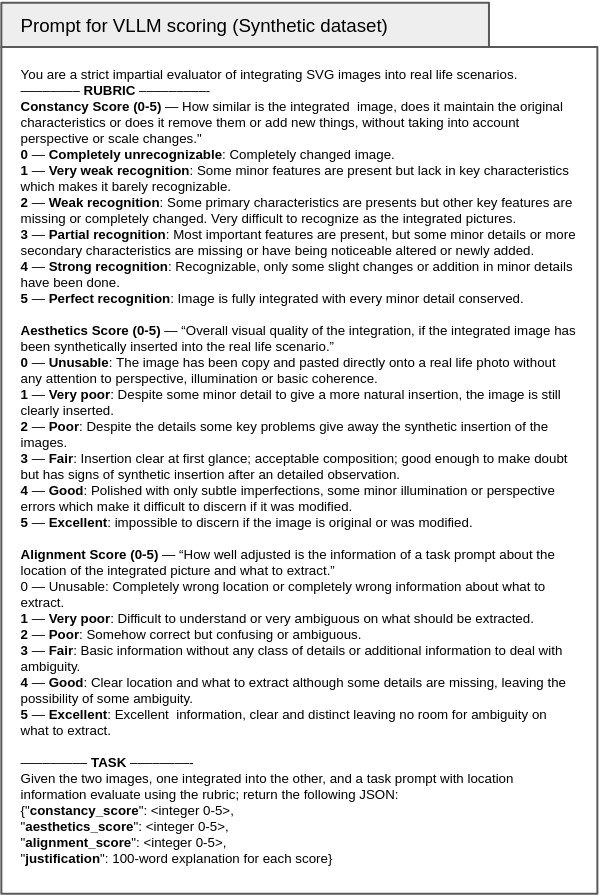}
        \caption{Prompt for scoring Synthetic WildSVG instances}
        \label{fig:prompt_judge_synthetic}
\end{figure}

We also present the prompts employed for VLM judging the instance of each dataset, \cref{fig:prompt_judge_natural} and \cref{fig:prompt_judge_synthetic}.\par

\subsubsection{Filtering score formulas}

The filtering procedure was performed using the following equations: \cref{formula:filter_score1} and \cref{formula:filter_score2}.\

\begin{equation}
\label{formula:filter_score1}
\begin{aligned}
Synthetic\ Dataset\  Score &= 40\% \cdot C_{\text{score}} + 40\% \cdot AE_{\text{score}} + 20\% \cdot AL_{\text{score}}, \\
\end{aligned}
\end{equation}
\begin{equation}
\label{formula:filter_score2}
\begin{aligned}
Natural\ Dataset\ Score &= 60\% \cdot C_{\text{score}} + 40\% \cdot AL_{\text{score}}.
\end{aligned}
\end{equation}
\subsubsection{Dataset statistics}
\label{sec:dataset_analysis}
\begin{figure}[htbp]
        \centering
        \includegraphics[width=1\linewidth]{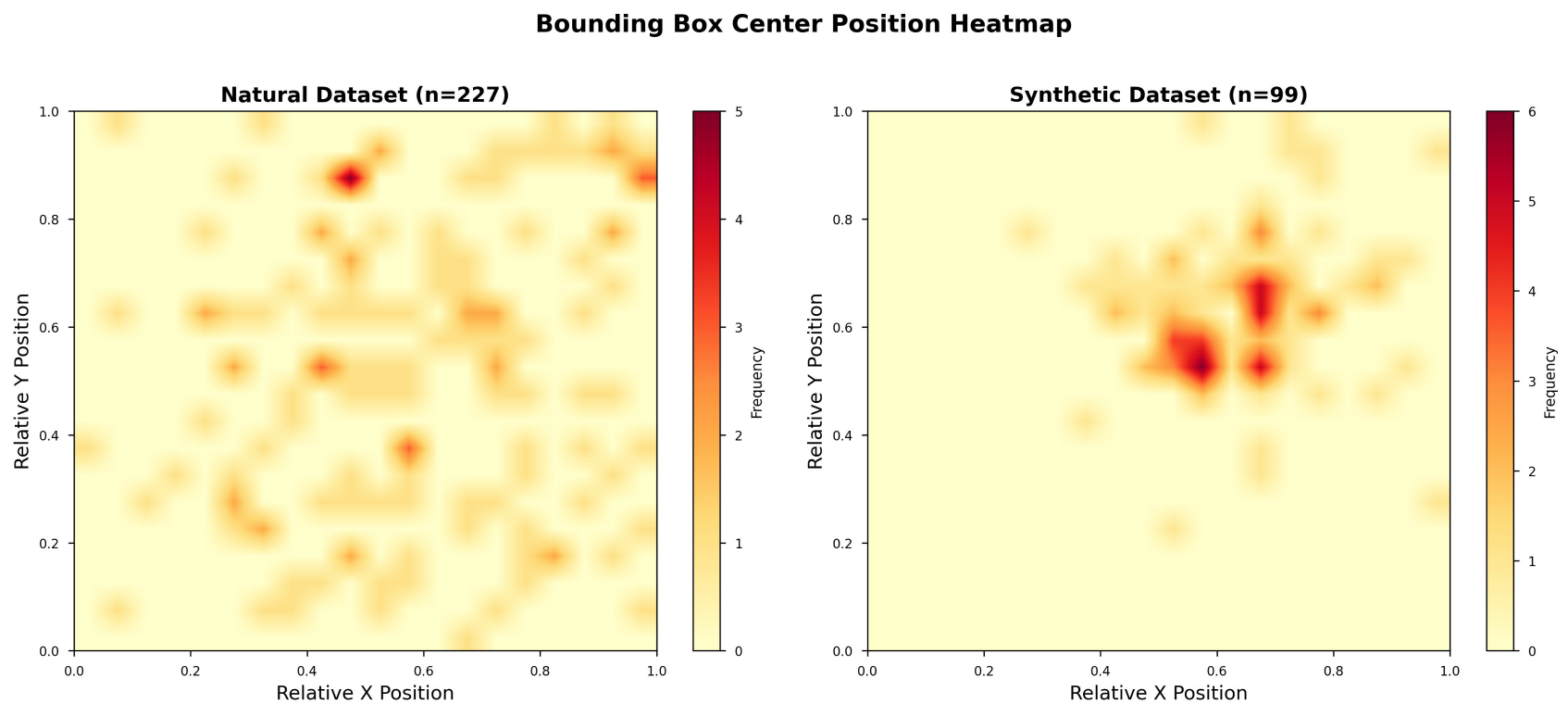}
        \caption{Bounding boxes position heatmap for WildSVG dataset test split}
        \label{fig:bbox_heatmap}
\end{figure}

As shown in \cref{fig:bbox_heatmap}, the heatmap reveals a more diverse spatial distribution of bounding boxes in the natural dataset, while the synthetic dataset exhibits a strong central bias. This observation motivated our decision to increase the weight of WilSVG for these detections, specifically to mitigate this type of positional bias.\par

\begin{figure}[htbp]
        \centering
        \includegraphics[width=1\linewidth]{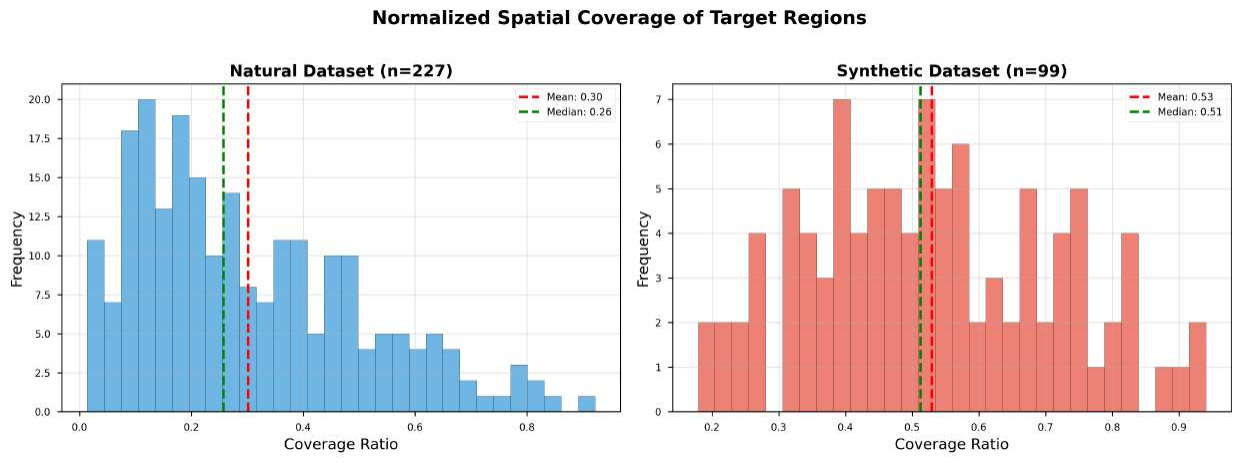}
        \caption{Distribution of bounding boxes coverage of the image for WildSVG dataset test split}
        \label{fig:bbox_distribution}
\end{figure}

From \cref{fig:bbox_distribution}, we observe that bounding boxes in the natural dataset tend to be smaller, with a clear skew toward reduced sizes. In contrast, the synthetic dataset displays a more uniform distribution of bounding box dimensions. In the synthetic case, the mean and median sizes are both around 0.5, which generally makes SVG detection easier. The combination of smaller bounding boxes and greater positional variability makes the natural dataset inherently more challenging, especially when real world noise is also considered.\par

\begin{figure}[htbp]
        \centering
        \includegraphics[width=1\linewidth]{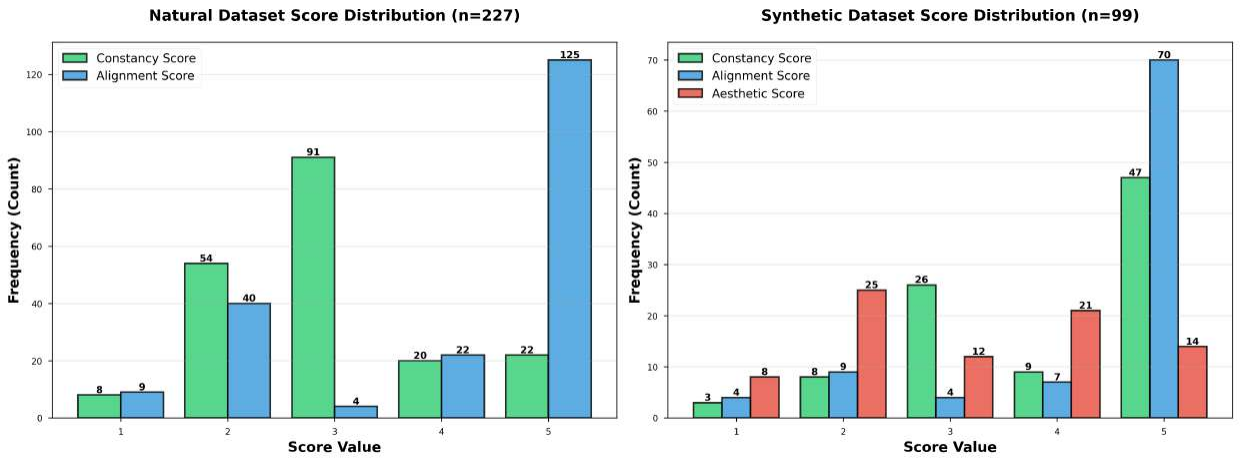}
        \caption{Score distribution for WildSVG dataset test split}
        \label{fig:score_distribution}
\end{figure}

Examining the score distributions in \cref{fig:score_distribution} reveals several additional points. Overall, both datasets exhibit reasonable alignment scores. However, the natural dataset shows slightly lower consistency compared with the synthetic one. This can be attributed to the fact that real world logos often contain small variations, so even when an SVG is conceptually similar, minor differences in color or spatial layout make exact matches harder to retrieve from the database. The aesthetic scores of the synthetic dataset also indicate that SVG integration can still be improved, since the distribution remains centered around a medium score of about 3 out of 5.\par

\subsubsection{Additional qualitative results on WildSVG}
\begin{figure*}[htbp]
        \centering
        \includegraphics[width=1\linewidth]{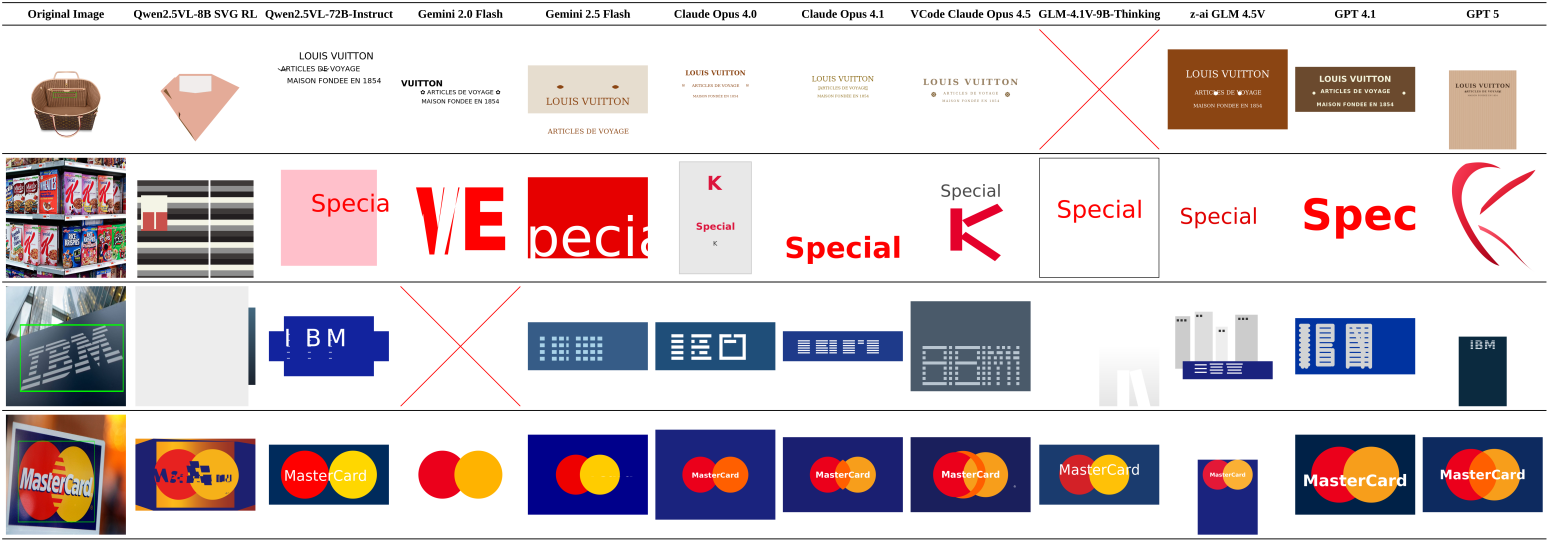}
        \caption{Comparison of VLMs for one-step SVG extraction natural dataset}
        \label{fig:complete_vllm_benchmark_quality_full_natural}
    \end{figure*}

\begin{figure*}[b]
        \centering
        \includegraphics[width=1\linewidth]{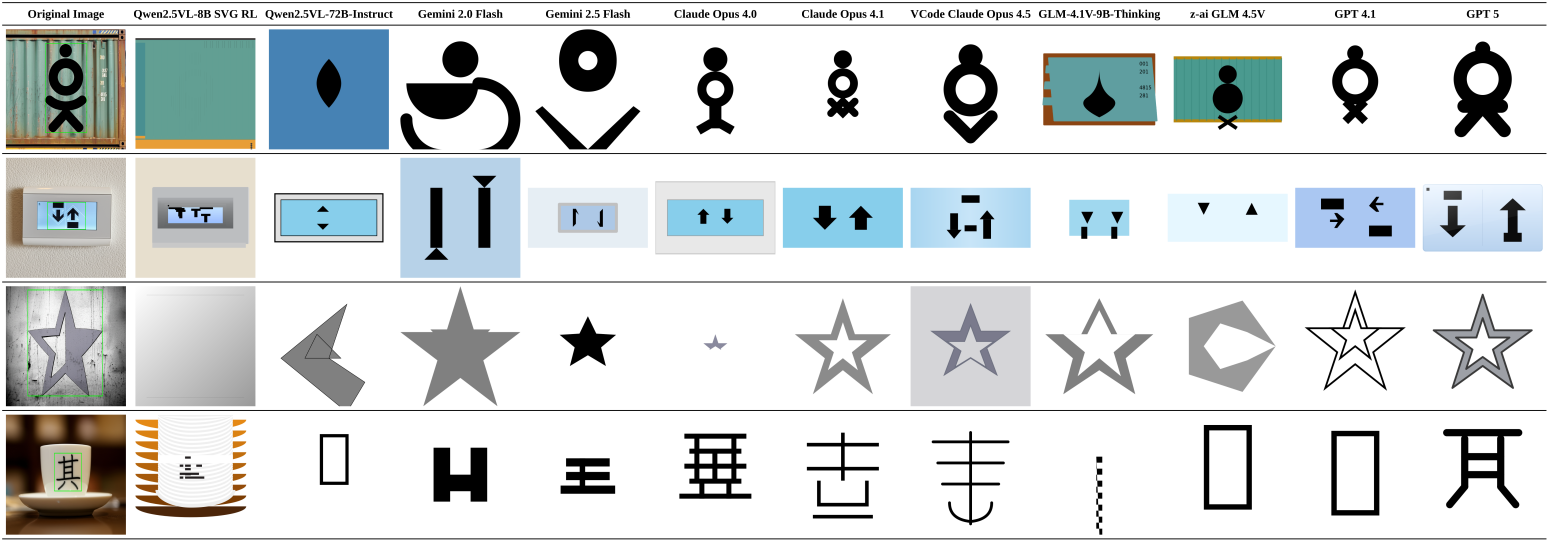}
        \caption{Comparison of VLMs for one-step SVG extraction synthetic dataset}
        \label{fig:complete_vllm_benchmark_quality_full_synthetic}
    \end{figure*}

\begin{figure*}[htbp]
        \centering
        \includegraphics[width=1\linewidth]{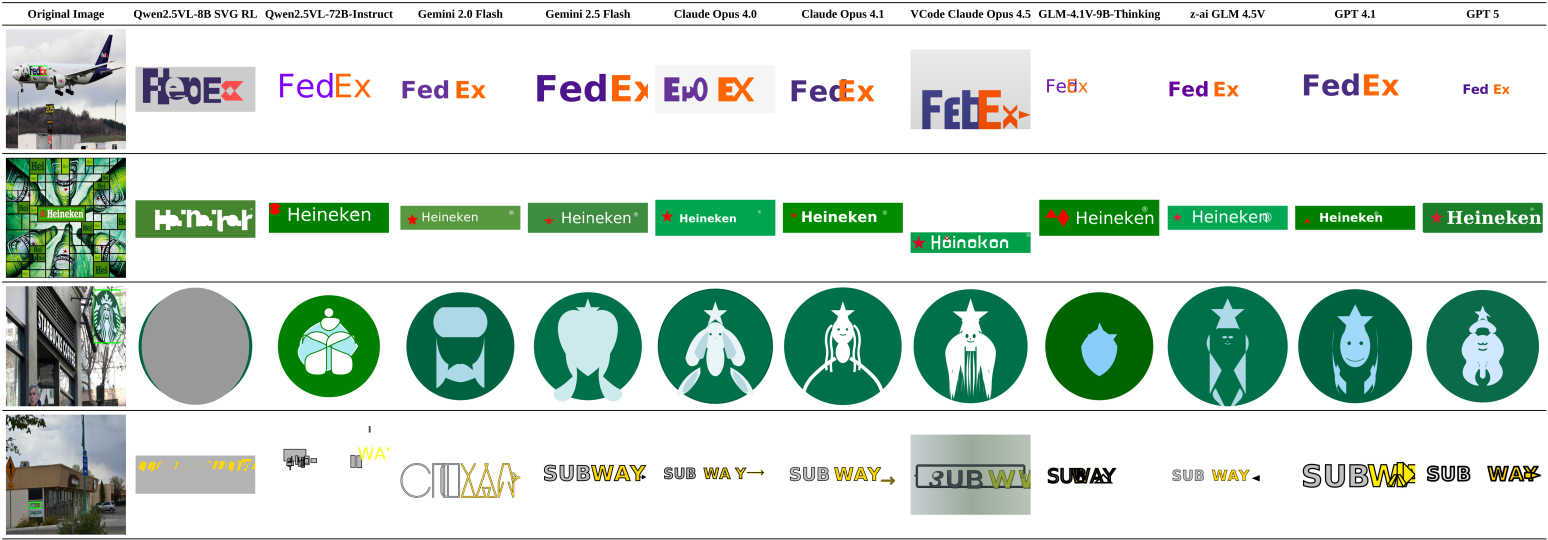}
        \caption{Comparison of VLMs for two-step SVG extraction natural dataset}
        \label{fig:complete_vllm_benchmark_quality_pp_natural}
    \end{figure*}

We present more results of the different SVG extraction approaches and models; \cref{fig:complete_vllm_benchmark_quality_full_natural}, \cref{fig:complete_vllm_benchmark_quality_full_synthetic} and \cref{fig:complete_vllm_benchmark_quality_pp_natural}.
    % First subfigure

    % Second subfigure
    
\subsubsection{WildSVG Dataset}
\begin{figure}[ht!]
        \centering
        \includegraphics[width=0.7\linewidth]{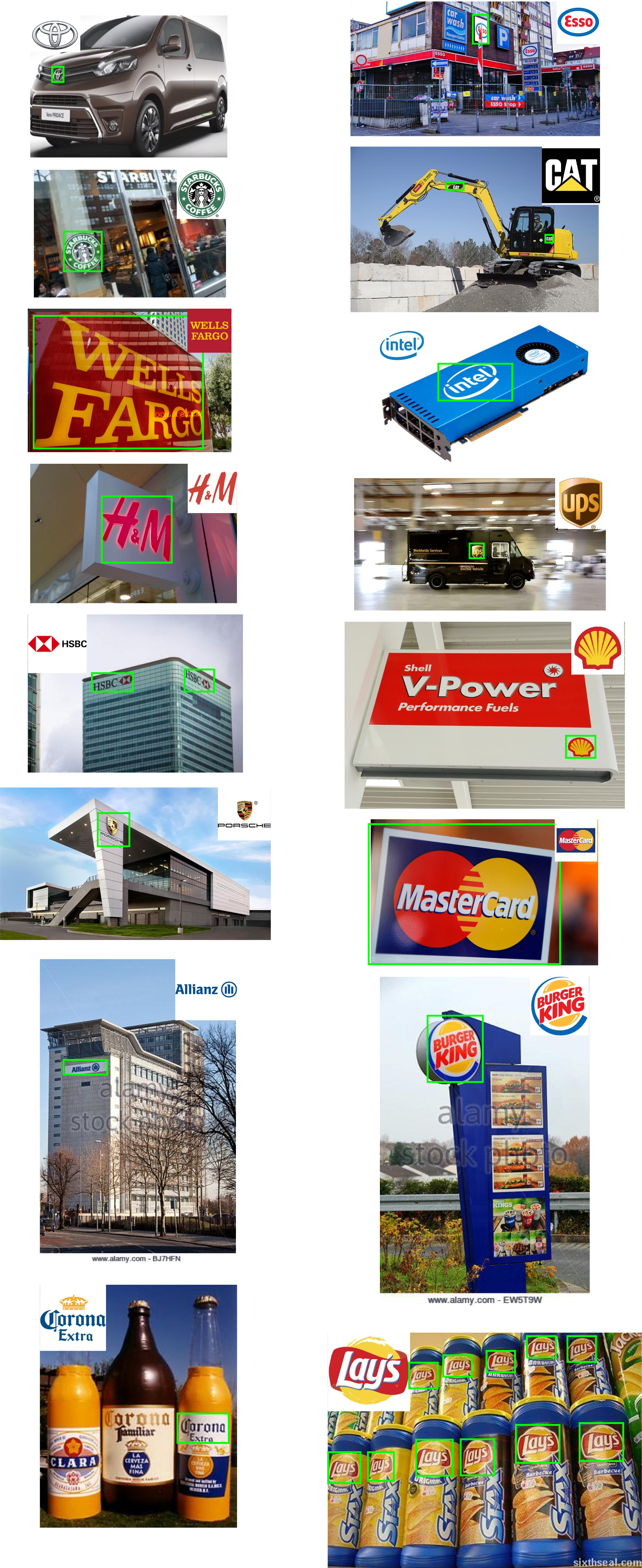}
        \caption{Examples of real-life images and associated SVG for natural WildSVG}
        \label{fig:full_natural}
\end{figure}

\begin{figure}[ht!]
        \centering
        \includegraphics[width=0.6\linewidth]{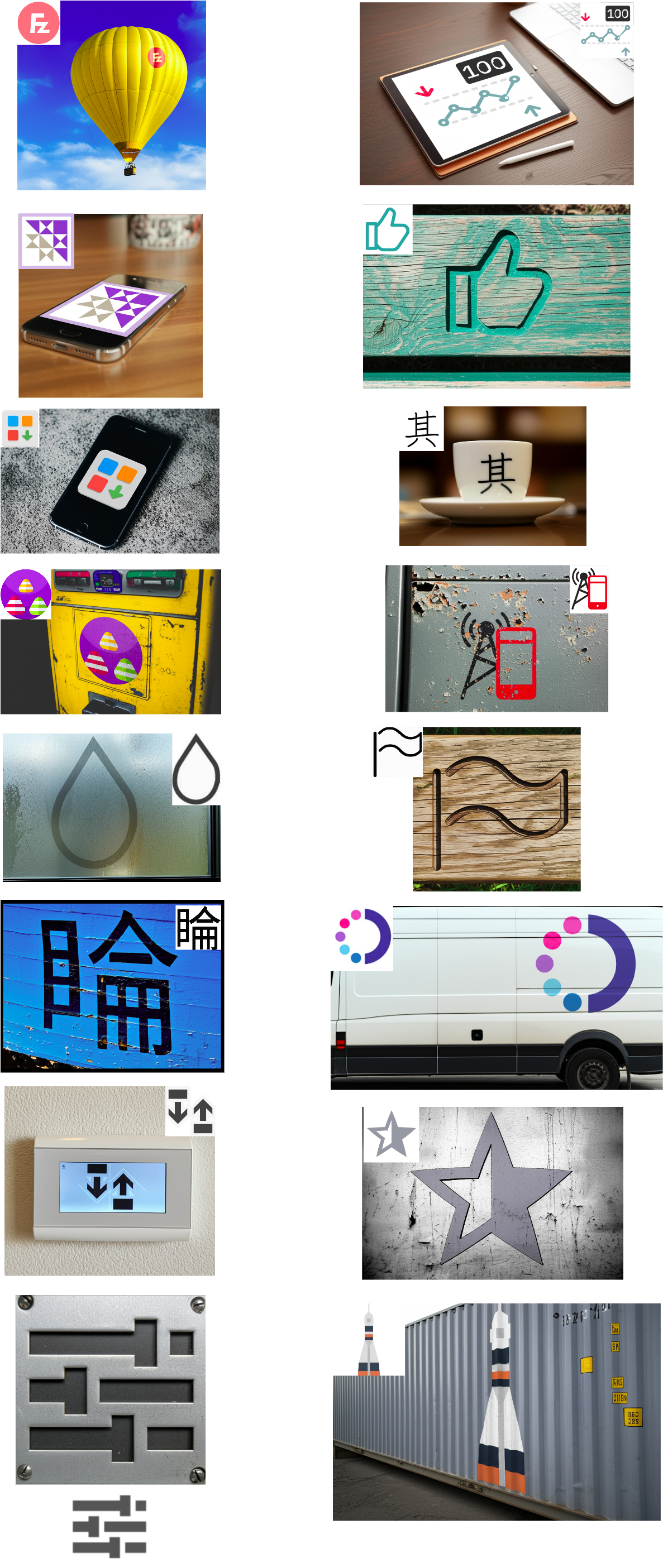}
        \caption{Examples of real-life images and associated SVG for synthetic WildSVG}
        \label{fig:full_synthetic}
\end{figure}

We present further examples from WildSVG dataset in \cref{fig:full_natural} and \cref{fig:full_synthetic}.

% \end{document}

\end{document}